\documentclass{article}

\usepackage[final,nonatbib]{nips_2017}

\usepackage[utf8]{inputenc} 
\usepackage[T1]{fontenc}    
\usepackage{hyperref}       
\usepackage{url}            
\usepackage{booktabs}       
\usepackage{amsfonts}       
\usepackage{nicefrac}       
\usepackage{microtype}      
\usepackage{amsmath}
\usepackage{graphicx}
\usepackage{algorithmic}
\usepackage{algorithm}
\usepackage{xcolor}
\numberwithin{algorithm}{section}
\usepackage[export]{adjustbox}

\title{Golfer: Trajectory Prediction with Masked Goal Conditioning MnM Network}

\author{
Xiaocheng Tang,
Soheil Sadeghi Eshkevari,
Haoyu Chen,
Weidan Wu,
Wei Qian,
Xiaoming Wang \\
DiDi Research, Autonomous Driving
}

\begin{document}

\maketitle

\begin{abstract}
Transformers have enabled breakthroughs in NLP and computer vision, and have recently began to show promising performance in trajectory prediction for Autonomous Vehicle (AV). How to efficiently model the interactive relationships between the ego agent and other road and dynamic objects remains challenging for the standard attention module. In this work we propose a general Transformer-like architectural module MnM network equipped with novel masked goal conditioning training procedures for AV trajectory prediction. The resulted model named \textbf{golfer} achieves state-of-the-art performance, winning the 2nd place in the 2022 Waymo Open Dataset Motion Prediction Challenge and ranked 1st place according to minADE.
\end{abstract}

\begin{figure}[h]
    \centering
    \includegraphics[scale=.47]{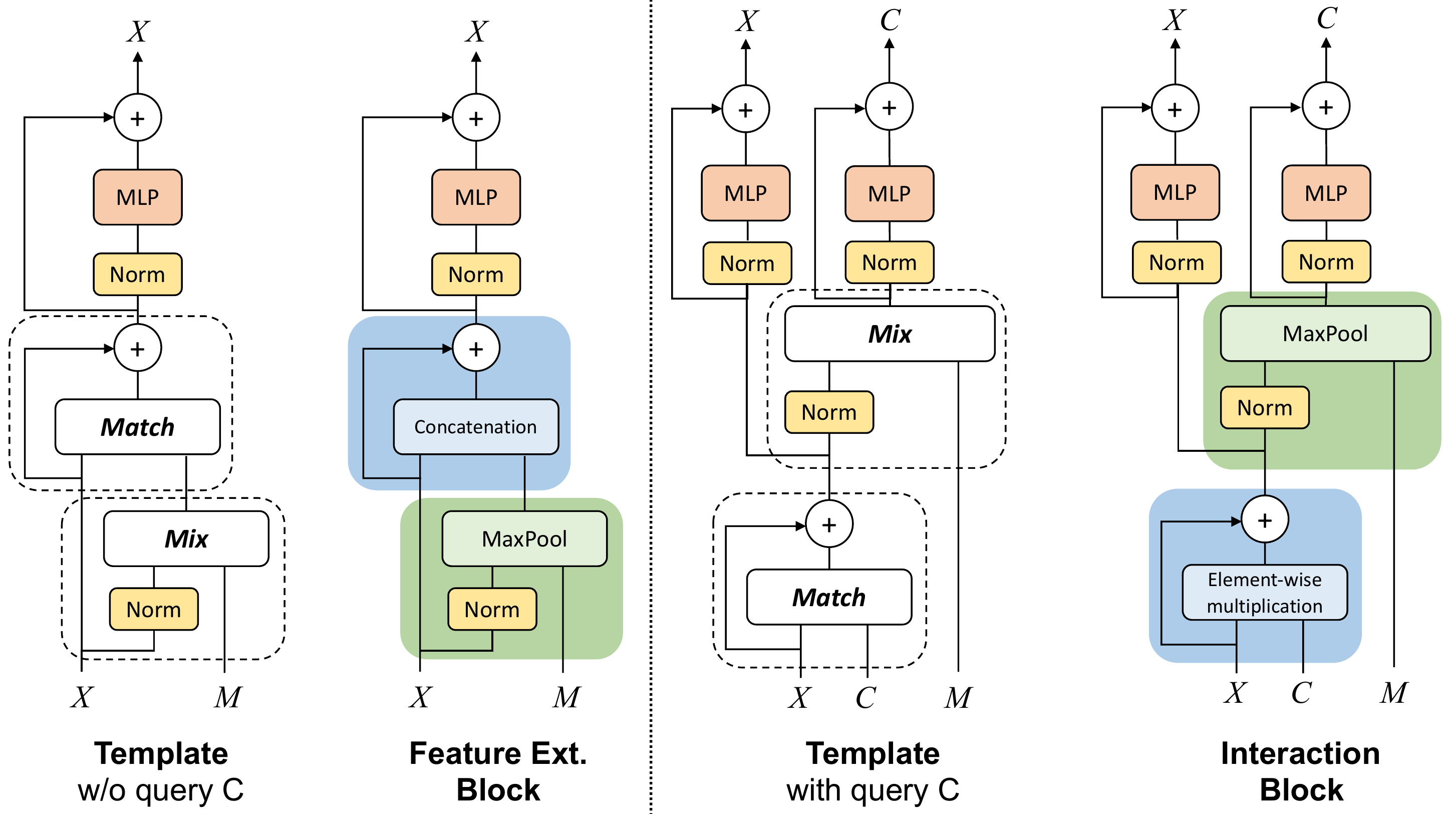}
    \caption{The general structure of \textbf{MnM Network} and the specific instantiation in our \textbf{golfer} model that was ranked \textbf{1st} according to minADE and \textbf{2nd} according to Soft mAP in the 2022 Waymo Open Dataset Motion Prediction challenge.}
    \label{fig:mnm}
\end{figure}

\section{Introduction}

Transformers have achieved remarkable success and gained much interest in multiple fields including NLP and computer vision. Its performance in trajectory prediction, however, remains unsatisfying \cite{strans_Ngiam:2021,Varadarajan:2021vn}. Particularly, trajectory prediction in an autonomous vehicle setting poses unique challenges to Transformers partly due to its inefficiency in obtaining a conditioned interactive encoding for sets of elements, such as modeling the relationships between the ego agent and the road objects.
Additionally, recent work have started to raise questions on core design choices of Transformers especially the attention component and whether it is essential to the remarkable effectiveness of Transformers \cite{Yu:2021tl,tolstikhin2021mlpmixer,NEURIPS2021_4cc05b35}. One of the very recent work \cite{tolstikhin2021mlpmixer}, for example, replaces the attention module with MLPs and still attains competitive performance on image classification benchmarks.
In fact, even earlier work such as PointNet \cite{qi2016pointnet} already shows that using a simple max pooling as the token mixers can achieve strong performance for 3D classification and segmentation.

Here we explore the effectiveness of Transformer-like architectures in AV trajectory prediction. Specifically, we propose a general architectural module, \textbf{Mix and Match} (MnM) block, that is especially effective at modeling conditioned interactive relationships and that encompasses Transformer as a special instance.
The resulted \textbf{MnM network}, consisted of only the stacked MnM blocks, is shown to achieve superior performance by directly predicting the trajectories from the raw point-like sets of agent and road inputs.
Particularly, equipped with the novel masked goal conditioning and the MnM network our trajectory prediction model named \textbf{golfer} ranked the second place on the Waymo Open Motion Dataset leaderboard as of May 23, 2022.

Waymo motion prediction competition has been a legacy platform for benchmarking best efforts for motion prediction in self-driving industry and research. In this competition, the objective is to predict agent trajectories for 8.0 seconds in the future and for at most eight heterogeneous agents in any given scene. The competition is an annual event that scores submissions based on a set of evaluation metrics including mAP, minADE, minFDE, miss rate, etc. In this report, we present an overview of the model that was ranked \textbf{1st} according to minADE and \textbf{2nd} according to Soft mAP in the 2022 Waymo motion prediction competition.

In summary, we propose a general form of set transformation that is found strongly effective in learning cross-correlations between items of a set as well as a set of items and a shared context. This transformation block is used in various settings throughout the model architecture. In terms of data encoding, we propose a hierarchical vector-based scene representation that enables a multi-level feature extraction scheme. To improve the efficacy of the predictive model, we use ensemble of models for the final prediction. More details on the main contributions of the designed model is given in the following sections.

\section{Method}
\label{sec:prob}

\subsection{MnM Block} 
\label{sub:mnm_network}

We present the core component of the golfer architecture, which we denote as the Mix and Match (MnM) Block. An illustration is shown in Figure \ref{fig:mnm}. It is the generalization of the transformers such that the attention mechanism can be formulated as a special instance of the Mix and Match operation we will introduce below.
We will see that with the abstraction of the MnM block we are able to replace the expensive attention computation with a much more efficient operation, as simple as a pooling layer, while retaining most expressive power, if not more, of the transformers.

Let the input be embedded as $X$ which consists of a sequence of $n$ tokens each with embedding dimension $d$. Let $M$ denote the binary mask of $X$ that for each token $X_i$ we have one bit $M_i$ indicating whether the corresponding token is valid or not. Additionally we define a query vector $C$ with the same embedding dimension as one token in $X$. The query vector $C$ is optional and depending on whether it is available the MnM block has two variants with minor differences, e.g., $X \leftarrow \text{MnM}(X, M)$ and $X, C \leftarrow \text{MnM}(X, C, M)$. The basic MnM block can be expressed as
\begin{align*}
    C &\leftarrow \text{Mix}(\text{Norm}(X), M), \\
    S &\leftarrow \text{Match}(C, X) + X, \\
    X &\leftarrow \sigma(\text{Norm}(S)W_1)W_2 + S.
\end{align*}
where $\text{Norm}(\cdot)$ denotes the normalization layer such as Batch or Layer Normalization and $\sigma(\cdot)$ is a non-linear activation function such as ReLU or GELU. Here we introduce two new operators, $\text{Mix}(\cdot, \cdot)$ and $\text{Match}(\cdot, \cdot)$.
\texttt{Mix} is a module that mixes token information and \texttt{Match} works by matching back the mixed information to each token.
As an example, consider the self-attention mechanism, where \texttt{Mix} is multiplying $X$ by itself followed by a softmax and \texttt{Match} takes the $n \times n$ attention matrix $C$ and multiplies it with $X$. Alternatively, and in a much simpler fashion, \texttt{Mix} can be simply a pooling operator, in which case $C$ will be a single vector, and \texttt{Match} can be either a concatenation or a element-wise product followed by a linear layer to map back to dimension $d$ if necessary. We argue that this is a more efficient alternative compared to the attention with no less expressive power.
The MnM block with a pre-given query vector $C$ is largely the same as the basic one but with an additional MLP layer on the query as follow
\begin{align*}
    S &\leftarrow \text{Match}(C, X) + X, \\
    C &\leftarrow \text{Mix}(\text{Norm}(S), M), \\
    X &\leftarrow \sigma(\text{Norm}(S)W_1)W_2 + S, \\
    C &\leftarrow \sigma(\text{Norm}(C)W_3)W_4 + C.
\end{align*}
Finally we note that a multi-headed version of MnM block can be obtained in a similar manner as in the transformers. The main advantage is parameter efficiency such that for the same size of the network we observe the best performance with Multi-headed MnM blocks.


\subsection{Auxiliary Masked Goal Conditioning} 
\label{sub:auxiliary_masked_goal_conditioning}
Inspired by recent advances in both language modeling and goal-conditioned RL, we apply a masking strategy to the targeted trajectory and use that as an additional input to the model. Note that the prediction task corresponds to using a fully masked input (100\%), and that a zero masked input becomes trivial to predict, e.g., copying from input to the output. Anything in between transforms the problem to estimating the conditional probability. In practice we use a 85\% mask. We further restrict each sample to contain at most one unmasked state in the targeted trajectory and remove that state from the loss computation.
To encode the masked targets we treat it either as one of the agent or the road objects, and add it to the set of agents or road objects with a coin toss. We find that this helps the network to capture the nonlinear mappings between the input and the output, e.g., a small change in the current scene sometimes demands a dramatic behavior correction.

As the training objective, we follow the common practice and use a double-headed loss with multi-mode Gaussian Mixture and a maximum likelihood classification head. The classification head predicts a softmax distribution over the mixture components. Each component is consisted of a time series of $(x, y)$ gaussian. We compare each component with the targeted trajectory and only compute the loss based on the one closest to the groundtruth in Euclidean distance.

\section{Golfer}\label{sec:golpher}

Our high-level architecture design consists of two main modules: Encoder and Decoder. The encoder module is designed to extract salient features from multi-model input channels (including maps and agents' past trajectories), encode interactions, and finally, produce a latent feature vector to the decoder module for final predictions. In motion planning task, the input data representation is a design choice that would affect the performance and efficacy. In this study, we propose a hierarchical scene and past trajectory representation that is concise and compatible with the MnM design. In this approach, roadmap objects as well as agents' histories are encoded as a sequence of points with certain embeddings. This representation shows multiple advantages: (1) extremely high data efficiency compared to the rasterized presentation, and (2) inherent capability of capturing semantic information compared to vector-based representations that encoded polylines into separate line segments with no context.

The encoder architecture includes three main steps: (a) feature extraction from agents' history, (b) feature extraction from roadmap polylines, (c) interaction modeling between extracted features and ego features. The outputs from last step are concatenated and passed to the decoder module for trajectory regression and classification.

The feature extractor design, as shown in Figure \ref{fig:mnm}, is inspired by graph neural networks (GNNs). Each input object (road polyline or agent history) is transposed into a sequence of multiple tokens (points in road polylines and agent coordinates in agent history) $S$ as well as a context vector $c$ that includes shared or global features for the group of tokens (for road polylines, it can be lane type and traffic signal encodings). In addition, binary mask $M$ indicates the availability of the sequential embeddings in case of encountering partial or incomplete poylines. Given these, the following feature extractor (FE) block is proposed based on the MnM block with:
\begin{align*}
    \text{Mix} \leftarrow \text{MaxPool} \hspace{4mm} \& \hspace{4mm}
    \text{Match} \leftarrow \text{Concat}.
\end{align*}
This block extracts nonlinear features from raw sequential embeddings, passes massages between sequential nodes via \textit{MaxPool(.)}, and incorporates global message via contextual vector $c$. For further improvement, a stacked and multi-headed version of the FE block is used in the final model.

In this design, the FE block extracts features from points of each polyline (local feature extraction). The polyline transformed features are then aggregated by a permutation-invariant transformation to produce a vector of latent information for the given poyline. In the next level, latent features from all polylines in the current scene interact particularly with the vector of latent features for the ego agent. The process of interaction is also modeled by a multi-headed MnM block. Here, the matching operation is an element-wise multiplication between the ego feature vector (as context) and latent features tensor. The mixing operation is a \textit{MaxPool(.)}.

The ego context $f_E$ interacts with latent features tensors from roadmap polylines and agent histories separately to produce $f_R$ and $f_A$. The final product of the encoder module is computed as: $f_{enc} = \text{MLP}(\text{Concat}[f_E, f_R, f_A])$. The decoder design is more straight-forward: the encoded feature $f_{enc}$ is passed through independent branches of \textit{MLP(.)}'s to predict a set of probable future trajectories along with corresponding occurrence probabilities.

Note that due to the multi-level process of feature extraction starting from point-level feature transformation to polyline-level feature aggregation and finally, scene-level feature transformation and aggregation, we term the encoder design as a "hierarchical scene encoding" that is compatible with heterogeneous data.

The final output is an ensemble of multiple models. In particular, we collect all outputs from the models and apply a weighted clustering using kmeans with a predefined number of clusters, e.g., 6. The centroid of each cluster and its corresponding aggregated and normalized weight is used as the predicted trajectory and the associated probability score.

\bibliographystyle{abbrv}
\bibliography{wodm}

\end{document}